\documentclass[sigconf]{acmart}

\renewcommand\footnotetextcopyrightpermission[1]{} 
\AtBeginDocument{%
  \providecommand\BibTeX{{%
    \normalfont B\kern-0.5em{\scshape i\kern-0.25em b}\kern-0.8em\TeX}}}

\acmYear{2023}
\setcopyright{acmlicensed}\acmConference[ICAIF '23]{4th ACM International
Conference on AI in Finance}{November 27--29, 2023}{Brooklyn, NY, USA}
\acmBooktitle{4th ACM International Conference on AI in Finance (ICAIF '23),
November 27--29, 2023, Brooklyn, NY, USA}
\acmPrice{15.00}
\acmDOI{10.1145/3604237.3626842}
\acmISBN{979-8-4007-0240-2/23/11}

\usepackage{algorithmic}
\usepackage{algorithm}

\usepackage{stfloats} 
\usepackage{multirow} 
\usepackage{acmart-taps}

\begin{document}

\title{From random-walks to graph-sprints: a low-latency node embedding framework on continuous-time dynamic graphs}

\author{Ahmad Naser Eddin}\authornote{These authors contributed equally to this work}
\orcid{0000-0002-6854-4800}

\affiliation{%
  \institution{Feedzai}
  \streetaddress{}
  \city{Coimbra}
  \state{}
  \country{Portugal}
  \postcode{}
}
\affiliation{%
  \institution{Departamento de Ciência de Computadores, Faculdade de Ciências, Universidade do Porto}
  \streetaddress{Rua do Campo Alegre 1021 1055}
  \city{Porto}
  \state{}
  \country{Portugal}
  \postcode{4169-007}
}
\email{ahmad.eddin@feedzai.com}

\author{Jacopo Bono}\authornotemark[1]
\affiliation{%
  \institution{Feedzai}
  \streetaddress{}
  \city{Coimbra}
  \state{}
  \country{Portugal}
  \postcode{}
}
\email{jacopo.bono@feedzai.com}

\author{David Apar\'icio}
\affiliation{%
  \institution{Departamento de Ciência de Computadores, Faculdade de Ciências, Universidade do Porto}
  \streetaddress{Rua do Campo Alegre 1021 1055}
  \city{Porto}
  \state{}
  \country{Portugal}
  \postcode{4169-007}
}
\email{daparicio@dcc.fc.up.pt}

\author{Hugo Ferreira}
\affiliation{%
  \institution{Feedzai}
  \streetaddress{}
  \city{Coimbra}
  \state{}
  \country{Portugal}
  \postcode{}
}
\email{hugo.ferreira@feedzai.com}

\author{Jo\~ao Ascens\~ao}
\affiliation{%
  \institution{Feedzai}
  \streetaddress{}
  \city{Coimbra}
  \state{}
  \country{Portugal}
  \postcode{}
}
\email{joao.ascensao@feedzai.com}

\author{Pedro Ribeiro}
\affiliation{%
  \institution{Departamento de Ciência de Computadores, Faculdade de Ciências, Universidade do Porto}
  \streetaddress{Rua do Campo Alegre 1021 1055}
  \city{Porto}
  \state{}
  \country{Portugal}
  \postcode{4169-007}
}
\email{pribeiro@dcc.fc.up.pt}

\author{Pedro Bizarro}
\affiliation{%
  \institution{Feedzai}
  \streetaddress{}
  \city{Coimbra}
  \state{}
  \country{Portugal}
  \postcode{}
}
\email{pedro.bizarro@feedzai.com}
\renewcommand{\shortauthors}{Eddin et al.}

\begin{abstract}
Many real-world datasets have an underlying dynamic graph structure, where entities and their interactions evolve over time. Machine learning models should consider these dynamics in order to harness their full potential in downstream tasks.\enlargethispage{12pt}
Previous approaches for graph representation learning have focused on either sampling k-hop neighborhoods, akin to breadth-first search, or random walks, akin to depth-first search. However, these methods are computationally expensive and unsuitable for real-time, low-latency inference on dynamic graphs. To overcome these limitations, we propose \textit{graph-sprints} a general purpose feature extraction framework for continuous-time-dynamic-graphs (CTDGs) that has low latency and is competitive with state-of-the-art, higher latency models. To achieve this, a streaming, low latency approximation to the random-walk based features is proposed. In our framework,
time-aware node embeddings summarizing multi-hop information are computed using only single-hop operations on the incoming edges.
We evaluate our proposed approach on three open-source datasets and two in-house datasets, and compare with three state-of-the-art algorithms (TGN-attn, TGN-ID, Jodie). We demonstrate that our \textit{graph-sprints} features, combined with a machine learning classifier, achieve competitive performance (outperforming all baselines for the node classification tasks in five datasets). Simultaneously, \textit{graph-sprints} significantly reduce inference latencies, achieving close to an order of magnitude speed-up in our experimental setting.\enlargethispage{12pt}
\aptLtoX[graphic=no,type=html]{}{\begin{figure}[htb]
\center
\includegraphics[width=0.45\textwidth]{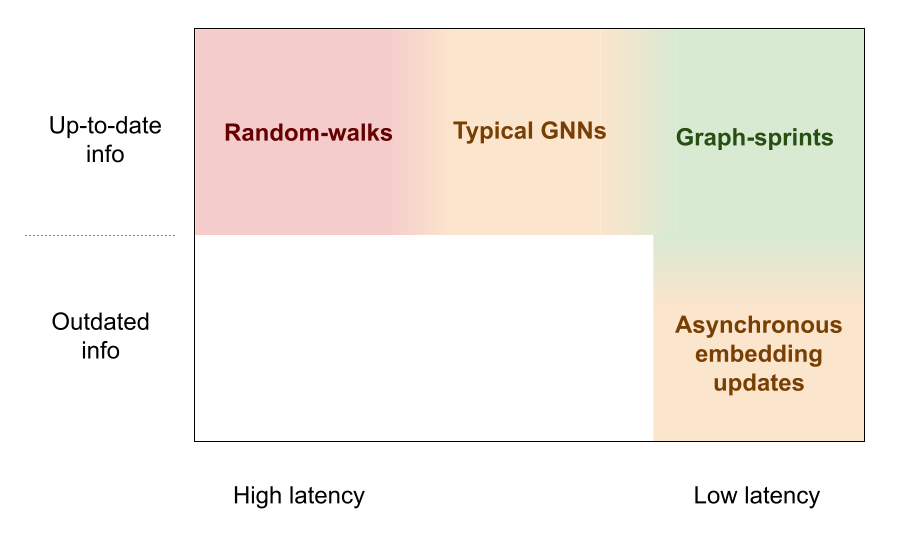}
\caption{\textbf{CTDGs Approaches Overview.} Current approaches either compute embeddings using real-time information but sacrificing latency, or compute low-latency embeddings but use outdated information in the computation. Our proposed method, \textit{graph-sprints}, computes embeddings in low latency using real-time information.}
\label{fig:solution_pipeline}
\end{figure}}
\textcolor{red}{This work has been published at ACM ICAIF 2023 conference proceedings.}
\end{abstract}


\aptLtoX[graphic=no,type=html]{\begin{figure}[htb]
\center
\includegraphics[width=0.45\textwidth]{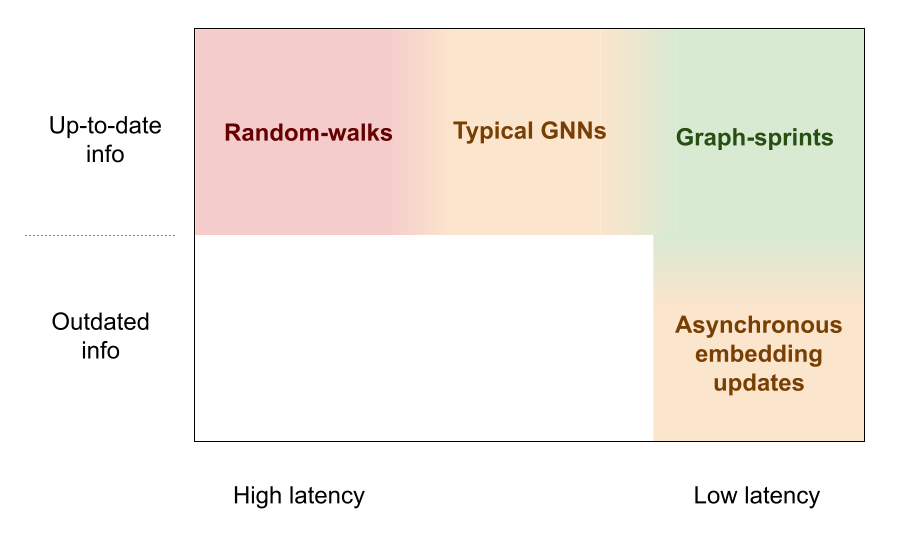}
\caption{\textbf{CTDGs Approaches Overview.} Current approaches either compute embeddings using real-time information but sacrificing latency, or compute low-latency embeddings but use outdated information in the computation. Our proposed method, \textit{graph-sprints}, computes embeddings in low latency using real-time information.}
\label{fig:solution_pipeline}
\end{figure}}{}
\maketitle

\begin{figure*}
\center
\includegraphics[width=0.88\textwidth]{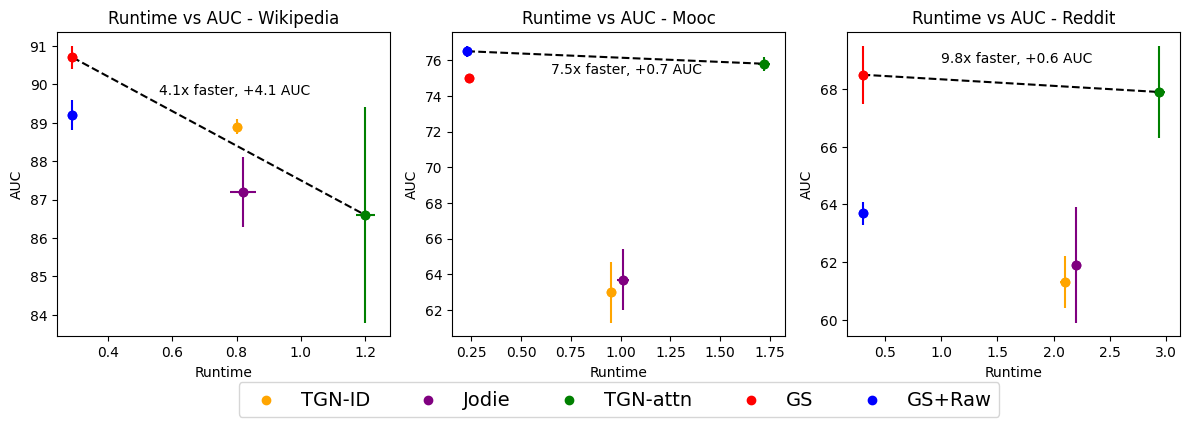}
\caption{\textbf{AUC vs. Runtime trade-off.} Our proposed methods based on \textit{graph-sprints} (GS or GS-Raw), allow for low latency inference while outperforming state-of-the-art methods in terms of AUC on node classification tasks. X-axis represents time in seconds to process 200 batches of size 200, Y-axis represents test AUC. Error bars denote the standard deviation over 10 random seeds.}
\label{fig:run_time_auc}
\end{figure*}

\section{Introduction}

Many real-world datasets have an underlying graph structure. In other words, they are characterized not only by their individual data points but also by the \textit{relationships} between them. Moreover, they are typically dynamic in nature, meaning that the entities and their interactions change over time. Examples of such systems are social networks, financial datasets, and biological systems~\cite{fan2019graph, wang2021review, zhang2021graph}. Dealing with dynamic graphs is more challenging compared to static graphs, especially if the graphs evolve in continuous time (also known as continuous-time dynamic graphs or CTDGs). The majority of machine learning models on graph datasets are based on graph neural networks (GNNs), achieving state-of-the-art performance~\cite{wu2020comprehensive}. A few deep neural network architectures emerged to deal with CTDGs in the past years~\cite{souza2022provably, tgn_icml_grl2020, jin2019node2bits, kumar2019predicting, xu2020inductive, guo2022continuous, jin2022neural}. One drawback of these approaches is that one either needs to sample k-hop neighborhoods to compute the embeddings (e.g. \cite{tgn_icml_grl2020}) or perform random-walks (e.g. \cite{jin2022neural}). Both cases are computationally costly, resulting in high inference latencies. One solution is to decouple the inference from the expensive graph computations, like in APAN~\cite{wang2021apan}. Performing the graph aggregations asynchronously results in inference using outdated information. However, in large data and high-frequency use-cases, such as our own focus on detecting fraud in financial transactions, there is a crucial need for low-latency solutions capable of processing up to 1000 transactions per second. Moreover, it is desirable for these solutions to leverage the latest information available in order to enhance detection capabilities.

In this work, we propose a graph feature extraction framework, which computes node embeddings in the form of features that characterize a node's neighborhood in dynamic graphs. Importantly, our proposed framework is designed for low-latency settings while still using the most up-to-date information during the embedding calculations. Since we derive our framework starting from random-walk based methods, we term our method \emph{graph-sprints}. The learned \emph{graph-sprints} features can then be used in any downstream system, for example in a machine learning model or a rule-based system. We show how the proposed \emph{graph-sprints} features, combined with a neural network classifier, are faster to run while not sacrificing in predictive performance compared with the higher-latency GNNs (Figure \ref{fig:run_time_auc}). 

Our \textbf{contributions} include: First, we propose \textit{graph-sprints}, a graph feature extraction framework for CTDGs. Next, we introduce two strategies to minimize the memory requirements of the generated features. Lastly, we benchmark the node embedding quality against state-of-the-art GNNs using five datasets and two tasks: node classification and link prediction.

The remainder of the paper is organized as follows. We first discuss the proposed \emph{graph-sprints} framework in  Section~\ref{sec:streaming}. In Section~\ref{sec:results}, we evaluate our framework using three open-source datasets from different domains, and two in-house datasets from the money laundering domain. We discuss related work in Section~\ref{sec:related_work} and in Section \ref{sec:conclusions} we put forward our main conclusions.

\section{Methods}
\label{sec:methods}

\subsection{Random-walk based features}
\label{sec:random-walk-based-features}
Before describing our \emph{graph-sprints} framework, we briefly summarize a random-walk based feature extraction framework. We will then derive efficient \emph{graph-sprints} computations in the next section. The random-walk based feature extraction framework requires the following steps, to generate a node embedding for a seed node.

\begin{enumerate}
\item \textbf{Select the seed node.} This selection depends on the use-case, and for CTDGs typically one considers entities involved in new activity, for instance if the change on the graph is adding a new edge between two nodes, then each of these two nodes could be a candidate for a seed node. 
\item \textbf{Perform random-walks starting from the seed nodes.} During the random-walks, relevant data such as node or edge features of the traversed path are collected. The type of random-walks influences what neighborhood is summarized in the extracted features. Walks can be (un)directed, biased, and/or temporal.
\item \textbf{Summarize collected data.} The data collected over walks is aggregated into a fixed set of features, characterizing each seed node's neighborhood. Examples of such aggregations are the average of encountered numerical node or edge features, the maximum of encountered out-degree, etc.
\end{enumerate}
The computation of these features is costly, because multiple random-walks need to be generated for each seed node. For CTDGs, one would have to compute such features each time an edge arrives. This is infeasible for high-frequency use-cases such as fraud detection in financial transactions, where a decision about a transaction needs to be made in a few milliseconds. In the next section, we derive an efficient approximation to the above random-walk based features.

\subsection{Graph-sprints: streaming graph features}
\label{sec:streaming}

\begin{figure}
\center
\includegraphics[width=0.45\textwidth]{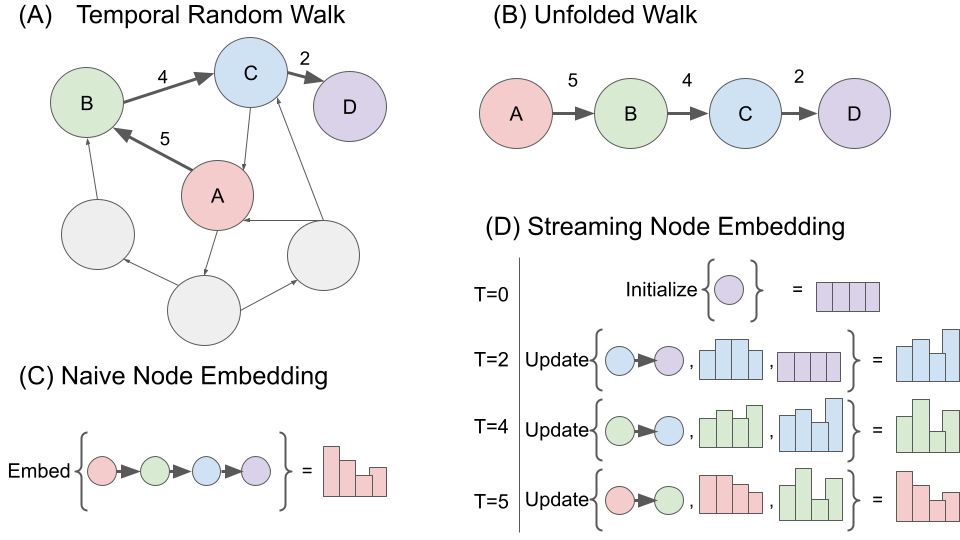}
\caption{\textbf{From random-walks to graph-sprints}. Edges have a timestamp feature (numbers) representing the time that a relationship was created. (A) A temporal random-walk is traversed from the most recent interaction A-B towards older interactions. (B) The same random-walk can be seen as a time-series of edges. (C) Based on a full temporal random-walk, one can compute embeddings by aggregating encountered feature values (section~\ref{sec:random-walk-based-features}). (D) One can compute similar embeddings in a streaming setting, from only the new edge and the existing embeddings of the involved nodes.}
\label{fig:streaming_walks}
\end{figure}

In this section, we propose approximations to random-walk based features described in Section~\ref{sec:random-walk-based-features}. Our aim in this section is to optimize the computation of such features by exploiting recurrence and abolishing the need to execute full random-walks (Figure \ref{fig:streaming_walks}). 

\subsubsection{Assumptions}

For our approximations to be reliable, we make the following assumptions: the input graph is a CTDG with directed edges (we will relax this assumption later), edges have timestamps and the temporal walks respect time, in the sense that the next explored edge is older than the current edge. With these assumptions, one can unfold any directed temporal walk as a time-series (Figure \ref{fig:streaming_walks}A and \ref{fig:streaming_walks}B). 

\subsubsection{Streaming histograms as node embeddings}

Given the above assumptions, we now formalize the approximation of random-walk based aggregations described in section \ref{sec:random-walk-based-features}. 

In this framework, we do not consider random-walks with a fixed number of hops, and instead consider infinite walks, on top of which we compute embeddings analogously to exponential moving averages. The importance of older information compared to newer is controlled by a factor $\alpha$ between $0$ and $1$. A larger $\alpha$ gives more weight to features further away in the walk (or in the past), and we can therefore consider $\alpha$ the parameter that replaces the number of hops. The ($1-\alpha$) factor is used to ensure the weights of the weighted average sum to 1.
Formally, let $\vec{s}_i$ be a histogram with $L$ bins, represented as an $L$-dimensional vector and characterizing the distribution of a feature $f$ in the neighborhood of node $i$. A full infinite walk starting at node 0 computes the histogram $\vec{s}_0$ as:  

\begin{equation}
\vec{s}_0 = \sum_{i=0}^{\infty} \alpha^{i} (1-\alpha) \vec{\delta}(f_i) \label{eq:infinite_sum}
\end{equation}

where $\sum$ is adding vectors, $\alpha$ is a discount factor between 0 and 1, controlling the importance of distant information in the summary $\vec{s}_0$, and $i$ denotes the hops of the walk ($i=0$ being the newest node, or in other words the seed node of the infinite walk). $f_i$ is the feature value at node $i$ or edge  $i$, and $\vec{\delta}(f_i)$ is an $L$-dimensional vector with element $\vec{\delta}_j=1$ if the feature value $f_i$ falls within bin $j$ and $\vec{\delta}_j=0$ for all other elements. Equation \ref{eq:infinite_sum} then implements a streaming counts per bin, where older information is gradually forgotten. If the feature $f_i$ is a node feature, then the value is taken from the current node. If it is an edge feature, then the feature value is taken from the edge connecting the current node and the chosen neighbor.

One could compute multiple such summaries per node, one for each node or edge feature of interest, and together they would summarize a neighborhood. The key idea is that we can now approximate the infinite random-walks, i.e., the infinite sum of equation~\ref{eq:infinite_sum}, by performing only a finite number of $k\geq 1$ hops, followed by choosing a random neighbor of the last encountered node and choosing an available summary $\vec{s}_{k}$ of that neighbor randomly, where $\vec{s}_{k}$ is defined as
\begin{equation}
\vec{s}_{k} = \sum_{i=0}^{\infty} \alpha^{i} (1-\alpha) \vec{\delta}(f_{i+k})
\end{equation}
With this strategy, we can approximate the summary $\vec{s}_0$ from equation~\ref{eq:infinite_sum} recurrently using 
\begin{equation}
\vec{s}_0 \approx   \sum_{i=0}^{k-1} \alpha^{i} (1-\alpha) \vec{\delta}(f_i) + \alpha^{k} \vec{s}_{k}\label{eq:truncated_sum} 
\end{equation}

Compared with equation \ref{eq:infinite_sum}, one now truncates the sum after $k$ terms. 
Note that whenever the last histogram $\vec{s}_{k}$ is normalized such that the bins sum to 1, e.g. using a uniform initialization for terminal nodes, equation \ref{eq:truncated_sum} guarantees that all subsequent histograms will be normalized in the same way. Since we are interested in low-latency methods, we take the limit of $k=1$ and Equation~\ref{eq:truncated_sum} becomes a streaming histogram:

\begin{equation}
\vec{s}_0 \leftarrow  (1-\alpha) \vec{\delta}(f_0) + \alpha \vec{s}_{1}\label{eq:exponentially_moving} 
\end{equation}

 The hyperparameter $\alpha$ can be chosen to depend on the number of hops or on time. When discounting by hops, this discount factor $\alpha$ is a fixed number between 0 and 1. When discounting by time, the factor is made dependent on the difference in edge timestamps, for example exponentially or hyperbolically.
 
Using equation \ref{eq:exponentially_moving}, one could approximate $N$ (biased) random-walks by sampling $N$ neighbors (non-uniformly), and subsequently combining the resulting histograms, e.g., by averaging. This would require performing $N$ 1-hop look-ups each time. 

Instead of that, we can increase efficiency even further by removing any stochasticity and updating a node's histogram at each edge arrival, combining the histograms of the two nodes involved in the arriving edge, as shown in equation~\ref{eq:graph_sprints}:

\begin{equation}
\vec{s}_0 \leftarrow  \beta \vec{s}_0 + (1-\beta) \left( (1-\alpha) \vec{\delta}(f_0) + \alpha \vec{s}_{1} \right)\label{eq:graph_sprints} 
\end{equation}

In this way we combine all neighbours' information implicitly using a moving average over time.

Hyperparameter $\beta$ is another discount factor between 0 and 1, controlling how much to focus on recent neighbor information in contrast to older information and which can optionally depend on time. In this way, we can update histograms in a fully streaming setting, using only information of each arriving edge. Nodes' summaries are initialized uniformly guaranteeing that the sum of all bins that belong to a certain edge is 1, e.g., if a feature has 10 bins then every bin is initialized with the value 0.1.
We term this procedure \emph{graph-sprints} and summarize it in algorithm~\ref{alg:streaming_singlesummary}. 

Compared to equation \ref{eq:exponentially_moving}, one can observe that the remaining sampling over single-hop neighbors is abolished, at the cost of imposing a more strict dependence on time. The advantage of algorithm~\ref{alg:streaming_singlesummary} is that no list of neighbors needs to be stored.   Moreover, algorithm~\ref{alg:streaming_singlesummary} can be applied in parallel to both the source node and the destination node, and therefore edges are not required to be directed. In fact, while we derived equation \ref{eq:graph_sprints} from random-walks, the attentive reader can notice that it can be interpreted as a special case of message passing where all neighbor summaries are aggregated using a weighted average, with weights that are biased by recency, and where the average is computed in a streaming fashion over time.

One special type of feature are the ones that have monotonically increasing values, exemplified by node degree features in a graph under conditions where edge deletion is non-existent. To avoid accumulating degrees over time, we propose to implement a streaming count of degrees per node. Every time an edge involving node $u$ arrives, we compute
\begin{equation}
    d_u = d_u \exp\left(-\Delta t / \tau_d\right) + 1 \label{eq:streaming_degrees}
\end{equation}
where $d_u$ denotes either in- or out-degree of node $u$, $\Delta t$ denotes the time differences between the current edge involving node $u$ and the previous one, and $\tau_d$ is a timescale for the streaming counts. This approach prevents values from growing too large and provides information about not only the structure but also the velocity and recency of edge additions. In financial crime detection, for example, it’s important to know if an account starts having more transactions than usual suddenly.

\begin{algorithm}
\caption{Graph-sprints (eq. \ref{eq:graph_sprints})}
\label{alg:streaming_singlesummary}

\begin{algorithmic}
\REQUIRE $EdgeStream$ \COMMENT{Stream of arriving edges $e_{i,j}$}
\REQUIRE $\mathcal{F}$ \COMMENT{Set of features for GS (e.g., node degree)}

\FOR{$e_{v,u} \in EdgeStream$}
  \STATE \textbf{Get} $\vec{s}_u, \vec{s}_v$ \COMMENT{Current summaries of nodes u,v}
  \STATE $\vec{s}^{\star}_v \gets \alpha \vec{s}_v$ \COMMENT{Multiply all bins by $\alpha$}

  \FOR{$f \in \mathcal{F}$}
    \IF{value($f$) in bin j}
      \STATE $\vec{s}^{\star}_{vj} \gets  \vec{s}^{\star}_{vj} + (1 - \alpha)$ \COMMENT{Add (1-$\alpha$) to bin $j$}
    \ENDIF
  \ENDFOR

  \STATE $\vec{s}_u \gets \beta \vec{s}_u + (1-\beta) \vec{s}^{\star}_v$ \COMMENT{Updated summary of node $u$.}
\ENDFOR
\end{algorithmic}
\end{algorithm}

\subsubsection{Choosing histogram bins}
Essential hyperparameters of this method are the choices of the boundaries of the histograms bins. We propose to use one bin per category for categorical features. If the cardinality of a certain feature is too high, we propose to form bins using groups of categories. For numerical features, one can plot the distribution in the training data and choose sensible bin edges, for example on every 10th percentile of the distribution. The framework is not constrained by one choice of bins, as long as they can be updated in a streaming way.

\subsection{Reducing space complexity}
\label{sec:reduce_storage}

The space complexity of the graph-sprints approach (algorithm~\ref{alg:streaming_singlesummary}) is \begin{equation}
M = \left| \mathcal{V} \right|  \sum_{f \in \mathcal{F}} L_f
\end{equation}
where $\left| \mathcal{V} \right|$ stands for the number of nodes, $L_f$ stands for the number of bins of the histogram for feature $f$, and $\mathcal{F}$ stands for the set of features chosen to collect in histograms. In case this memory is too high, we propose the following methods to reduce memory further.  

\textbf{Reducing histogram size using similarity hashing}
Following the similarity hashing approach proposed in \citet{jin2019node2bits}, we extend the method to the streaming setting. All histograms as defined in the previous sections are normalized (in the sense that bin values sum to 1), and we can concatenate them into one vector $\vec{s}_{tot}$. We can now define a hash mapping by choosing $k$ random hyperplanes in $\mathbb{R}^M$ defined by unit vectors ${\vec{h}_j}, j=1,\dots ,k$.

The inner product between the histograms vector and the $k$ unit vectors results in a vector of $k$ values, each value $\theta$ can be calculated using the dot product of the unit vector $\vec{h}_j$ and the histogram vector $\vec{s}_{tot}$, as illustrated in Equation~\ref{eq:real_hash}. We use the superscript $t$ to denote the current time step.

\begin{equation}
\theta_{j}^t = \vec{h}_j\cdot \vec{s}_{tot}^t \label{eq:real_hash}
\end{equation}

One can binarize the representation of the hashed vector using by taking the \emph{sign} of the above $\theta_{j}^t$.

Therefore, the resulting space complexity per node is $k$, replacing the number of bins in the memory $M$ by the number of hash vectors $k$. 

Importantly, the hashed histograms can be updated without storing any of the original histograms. Combining equations \ref{eq:exponentially_moving} and equation~\ref{eq:real_hash} and denoting $\vec{\delta}(\vec{f})$ the concatenation of the $\vec{\delta}$ vectors for all collected features, we get
\begin{equation}
\theta_{j}^{t+1} = \theta_{j}^t \cdot \alpha + \vec{h}_j\cdot \vec{\delta}(\vec{f})\cdot (1-\alpha)  \label{eq:update_theta}
\end{equation}

Therefore, we can compute the next hash $\theta_{j}^{t+1}$ or $\text{sign}(\theta_{j}^{t+1})$ directly from the previous $\theta_{j}^{t}$ and the new incoming features $\vec{\delta}(\vec{f})$.It is also important to note that this hashing scheme is preserved when averaging.

\textbf{Reducing embedding size using feature importance}
One can reduce the needed memory by relying on feature importance techniques. One possibility is to train a classifier on the raw node and/or edge features and determine feature importances, after which only the top important features are used in the \textit{graph-sprints} framework. Or similarly train on all bins and decide the bins to be used based on their importance in the classification task. Thus, either reducing the number of features, or the number of bins within the features, or both.

\section{Experiments}
\label{sec:results}

\subsection{Experimental setup}
We assess the quality of the graph based features generated by the \textbf{graph-sprints} framework on two different tasks, namely, node classification and link prediction.
 We use three publicly available datasets from the social and education domains and two proprietary datasets from money laundering domain. We detail their main characteristics in Table~\ref{table:external_data_stats} and Table~\ref{table:internal_data_stats}, respectively. All datasets are CTDGs and are labeled. Each dataset is split into train, validation, and test sets respecting time (i.e., all events in the train are older than the events in validation, and all events in validation are older than the events in the test set).
We use Optuna~\cite{akiba2019optuna} to optimize the hyperparameters of all models, training 100 models using the TPE sampler and with 40 warmup trials. Each model trains using early stopping with a patience of 10 epochs, where the early stopping metric computed on the validation set as area under ROC curve (AUC) for node classification and average precision (AP) for link prediction. All models were trained using a batch size of 200 edges. Hyperparameter ranges used during optimization are reported in Table~\ref{table:hpt_params}, and the optimized values are reported in Table~\ref{table:hpt_params_used}.

\begin{table}[b]
\centering
\caption{Information about public data~\cite{kumar2019predicting}. We adopt the identical data partitioning strategy employed by the baseline methods we compare against, which also utilized these datasets.}
\begin{tabular}{|c|c|c|c|}
\hline
\textbf{}  & \textbf{Wikipedia} & \textbf{Mooc} & \textbf{Reddit} \\ \hline
\#Nodes & 9,227 & 7,047 & 10,984 \\ \hline
\#Edges & 157,474 &  411,749 & 672,447 \\ \hline
Label type & editing ban &  student drop-out & posting ban \\ \hline
Positive labels & 0.14\% &  0.98\%  & 0.05\% \\ \hline
Used split & 75\%-15\%-15\% & 60\%-20\%-20\% & 75\%-15\%-15\% \\ \hline
\end{tabular}
\label{table:external_data_stats}

\end{table}

\subsubsection{Baselines}
\label{sec:baselines}

As a first baseline, we reproduce a state-of-the-art GNN model for CTDGs, the temporal-graph network (TGN)~\cite{tgn_icml_grl2020}, which leverages a combination of memory modules and graph-based operators to obtain node representations. 
As an important note, we mention that the pytorch geometric~\cite{Fey/Lenssen/2019} implementation of TGN was used, for which the sampling of neighbors uses a different strategy than the original TGN implementation. Indeed, the original paper allowed to sample from interactions within the same batch as long as they are older, while the pytorch geometric implementation does not allow within-batch information to be used. As also noted in the pytorch geometric documentation, we believe the latter to be more realistic. As a consequence, our TGN results are not directly comparable with the originally published TGN performances. In any case, the graph-sprints embeddings were computed using the same batch size and therefore also do not have access to within-batch information, allowing a fair comparison between the algorithms.

Two variations of the TGN architecture were used. First, TGN-attn was implemented, which was the most powerful variation in the original paper but is expected to be slower due to the graph-attention operations. Second, TGN-ID was implemented, which is a variation of the TGN where no graph-embedding operators are used, and only the embedding resulting from the memory module is passed to the classification layers. 

A third baseline we use is Jodie~\cite{kumar2019predicting}. We use the TGN implementatin of Jodie, where instead of using Graph attention embeddings on top of the memory embedding, a time projection embedding module is used and where the loss function is otherwise identical to the TGN setting. For a fair comparison with TGN we use the same memory updater module, namely, gated recurrent units (GRUs).

The TGN-ID and Jodie baselines do not require sampling of neighbors, and were therefore chosen as lower-latency baselines compared to TGN-attn.

\subsubsection{Graph-sprints and classifier}

For each arriving edge, we apply the graph-sprints feature update (algorithm \ref{alg:streaming_singlesummary}) to both the source node and the destination node in parallel. All edge features are used for the computation of the graph-sprints features, and for each feature bin edges are chosen as the 10 quantiles computed on the training data. Since the graph-sprints framework only creates features, a classifier is implemented for the classification tasks. We chose to implement a neural network consisting of dense layers, normalization layers, and skip-connections across every two dense layers. Hyperparameter optimization proceeds in two steps. First, default parameters for the classifier are used to optimize the discount factors of the \emph{graph-sprints} framework, $\alpha$ and $\beta$. For this step, 50 models are trained. Subsequently, hyperparameter optimization of the classifier follows same approach as TGN, training 100 models. 

In all experiments, we test the following three cases. Firstly, we train the classifier using only raw features (Raw). We then train the classifier using only the graph-sprint features (GS). Finally, we train the classifier using both raw and graph-sprint features (GS+Raw).

\subsubsection{Node Classification vs Link Prediction}

For the node classification task on the Wikipedia, Reddit and Mooc datasets, we concatenate the source and destination node embeddings and feed the concatenated vector to the classifier, as is usual for these datasets since labels are on the edge level, for instance, in Mooc dataset where edges connect students to courses, labels indicating whether the student will drop-out are on the edge level. 

For the link prediction task, all existing edges are considered positive edges, and negative edges are generated following the same approach as the original TGN paper \cite{tgn_icml_grl2020}, a negative edge is sampled for every positive one. 
We perform the link prediction task both in the \textit{transductive} and \textit{inductive} settings. In the \textit{transductive} setting, negative edges are sampled on the same graph used for training. Conversely, in the \textit{inductive} setting, the sampled negative edges are constrained to include at least one new node which was not used in the training graph.

\label{app:hp_ranges}

\begin{table}[htb]
\centering
\caption{Hyperparameters ranges for the Graph-Sprints (GS) method and the other baselines (GNN)}
\begin{tabular}{|c|c|c|c|}
\hline
\textbf{Method} & \textbf{Hyperparameter} & \textbf{min} & \textbf{max} \\ \hline
\textbf{GS} & $\alpha$ & 0.1 & 1  \\ \hline
\textbf{GS} & $\beta$  & 0.1 & 1 \\ \hline

\textbf{GNN/GS}  & Learning rate &  0.0001 & 0.01 \\ \hline
\textbf{GNN/GS}  & Dropout perc & 0.1 & 0.3 \\ \hline
\textbf{GNN/GS}  & Weight decay & 0.000000001 & 0.001 \\ \hline
\textbf{GNN/GS}  & Num of dense layers & 1 & 3 \\ \hline
\textbf{GNN/GS}  & Size of dense layer &  32 & 256 \\ \hline
\textbf{GNN/GS}  & Batch size & 128 & 2024 \\ \hline
\textbf{GNN}  & Memory size & 32 & 256 \\ \hline
\textbf{GNN}  & Neighbors per node & 5 & 10 \\ \hline
\textbf{GNN}  & Num GNN layers & 1 & 3 \\ \hline
\textbf{GNN}  & Size GNN layer & 32 & 256 \\ \hline

\end{tabular}
\label{table:hpt_params}

\end{table}

\subsection{Public datasets experiments} 

\subsubsection{Task performance}

In Table~\ref{table:results_external_data_nc} we report the average test AUC  ± std for the \textbf{Node classification} task. Our approach involved retraining the best model obtained after hyperparameter optimization, using 10 different random seeds. We can observe that on all datasets, the best model for node classification uses a variation of our graph-sprint framework (either GS or GS+Raw). In table~\ref{table:results_external_data_link_pred} we report the average test AUC ± std, along with the average precision (AP) ± std for the \textbf{Link prediction} task. Results were again computed after retraining the best model obtained through hyperparameter optimization, utilizing 10 distinct random seeds. We can observe that the graph-sprints model is the best for link prediction on the Mooc dataset, but performs poorly on the Wikipedia dataset. On the Reddit dataset, the graph-sprints model is second in the transductive setting, but is considerably worse for the inductive setting.

\begin{table}[b]
\centering
\caption{Results on Node classification task, we report the average test AUC ± std achieved by retraining the best model after hyperparameter optimization using 10 random seeds. Our models, Raw, GS, and GS+Raw, use the same ML classifier but differ in the features employed for training. Raw uses raw edge features, GS uses graph-sprints histograms, and GS+Raw combines both. We identify the \textcolor{red}{\textbf{best}} model and highlight the \textcolor{blue}{\underline{second best}} model.}
\begin{tabular}{|c|c|c|c|}
\hline
\multirow{2}{*}{\textbf{Method}} &\multicolumn{3}{|c|}{\textbf{AUC ± std}} \\ \cline{2-4}

  & \textbf{Wikipedia} & \textbf{Mooc} & \textbf{Reddit} \\ \hline
 
 Raw & 58.5 ± 2.2 & 62.8 ± 0.9 & 55.3 ± 0.8 \\ \hline

 TGN-ID  & 88.9 ± 0.2 & 63.0 ± 17 & 61.3 ± 2.0 \\ \hline
 Jodie   & 87.2 ± 0.9 & 63.7 ± 16.7 & 61.9 ± 2.0 \\ \hline
 
 TGN-attn & 86.6 ± 2.8 & \textcolor{blue}{\underline{75.8 ± 0.4}} & \textcolor{blue}{\underline{67.9 ± 1.6}} \\ \hline
 
 GS & \textcolor{red}{\textbf{90.7 ± 0.3}} & 75.0 ± 0.2 & \textcolor{red}{\textbf{68.5 ± 1.0}}  \\ \hline 

 GS+Raw  & \textcolor{blue}{\underline{89.2 ± 0.4}} & \ \textcolor{red}{\textbf{76.5 ± 0.3}} & 63.7 ± 0.4  \\ \hline 

\end{tabular}
\label{table:results_external_data_nc}
\end{table}

\begin{table*}[htb]
\centering
\caption{\textbf{Results on Link prediction task}, we report test average AUC and average precision (AP) ± std resulting from retraining the best model after hyperparameter optimization using 10 random seeds. We report results on both Transductive (T) or Inductive (I) settings. Our models, GS, and GS+Raw, use the same ML classifier but differ in the features employed for training. GS uses GS histograms, and GS+Raw combines GS histograms with raw edge features. We identify the \textcolor{red}{\textbf{best}} model and highlight the \textcolor{blue}{\underline{second best}} model.}
\begin{tabular}{|c|c|c|c|c|c|c|c|}
\hline
\multirow{2}{*}{} & \multirow{2}{*}{\textbf{Method}}& \multicolumn{2}{c|}{\textbf{Wikipedia}} & \multicolumn{2}{c|}{\textbf{Mooc}}  & \multicolumn{2}{c|}{\textbf{Reddit}} \\ \cline{3-8}
 & & {\textbf{AUC}} & \textbf{AP} & {\textbf{AUC}} & \textbf{AP} & {\textbf{AUC}} & \textbf{AP} \\ \hline

 \multirow{5}{*}{T}
   & \textbf{TGN-ID}  & \textcolor{blue}{\underline{95.6 ± 0.2}} & \textcolor{blue}{\underline{95.8 ± 0.1}} & 80.4 ± 5.8 & 75.0 ± 6.1 & 94.7 ± 0.7 & 93.2 ± 1.0\\ \cline{2-8}

 & \textbf{Jodie} & 94.3 ± 0.3 & 94.5 ± 0.3 & 
\textcolor{blue}{\underline{85.1 ± 1.8}} & \textcolor{blue}{\underline{80.0 ± 3.5}} & 94.9 ± 1.2 & 93.4 ± 1.7  \\ \cline{2-8}
 
 & \textbf{TGN-attn}  & \textcolor{red}{\textbf{ 97.0 ± 0.3}} & \textcolor{red}{\textbf{97.3 ± 0.3}} & 80.3 ± 8.1 & 75.6 ± 8.4 & 96.1 ± 0.4 & 95.1 ± 0.7\\ \cline{2-8}

& \textbf{GS} & 92.5 ± 0.6 & 92.9 ± 0.7 & 82.7 ± 0.8 & 81.1 ± 0.7 & \textcolor{blue}{\underline{96.1 ± 0.2}} & \textcolor{blue}{\underline{95.3 ± 0.3}} \\ \cline{2-8}

 & \textbf{GS+Raw}  & 92.1 ± 0.4 & 92.6 ± 0.4 & \textcolor{red}{\textbf{85.4 ± 0.3}} & \textcolor{red}{\textbf{83.7 ± 0.3}} & \textcolor{red}{\textbf{96.8 ± 0.1}} & \textcolor{red}{\textbf{96.1 ± 0.2}} \\ \hline \hline

 \multirow{5}{*}{I}
  & \textbf{TGN-ID}  & \textcolor{blue}{\underline{92.2 ± 0.2}} & \textcolor{blue}{\underline{92.8 ± 0.2}} & 68.5 ± 8.6 & 63.5 ± 6.7 & 93.4 ± 0.6 & 92.2 ± 0.8 \\ \cline{2-8}

 & \textbf{Jodie} & 87.0 ± 0.6 & 89.1 ± 0.7 & 
 71.1 ± 2.2 & 66.1 ± 2.9 & 92.3 ± 1.3 & 90.8 ± 1.9 \\ \cline{2-8}

 & \textbf{TGN-attn} & \textcolor{red}{\textbf{94.5 ± 0.2}}	& \textcolor{red}{\textbf{95.0 ± 0.2}} & 71.4 ± 4.1	& 66.9 ± 3.9  & \textcolor{red}{\textbf{95.0 ± 0.4}} & \textcolor{red}{\textbf{94.3 ± 0.5}} \\ \cline{2-8}

 & \textbf{GS} & 92.0 ± 0.3 & 91.7 ± 0.4 & \textcolor{blue}{\underline{78.2 ± 0.6}}	& \textcolor{blue}{\underline{76.5 ± 0.6}} & 92.7 ± 0.5 & 92.7 ± 0.6  \\ \cline{2-8}

 & \textbf{GS+Raw} & 91.4 ± 0.2 & 91.1 ± 0.3 & \textcolor{red}{\textbf{83.0 ± 0.5}} & \textcolor{red}{\textbf{80.3 ± 0.5}} & \textcolor{blue}{\underline{93.5 ± 0.4}} & \textcolor{blue}{\underline{92.2 ± 0.5}}  \\ \hline

\end{tabular}
\label{table:results_external_data_link_pred}
\end{table*}

\begin{table}[htb]
\centering
\caption{Hyperparameters used in node classification (NC) and link prediction (LP). Learning rate and Weight decay are approximated due to space constraints}
\begin{tabular}{|c|c|c|c|c|}
\hline
\textbf{Task} & \textbf{Hyperparameter} & \textbf{Wikipedia} & \textbf{Mooc} & \textbf{Reddit} \\ \hline
{\multirow{8}{*}{NC}} & \(\alpha\) & 0.7 & 0.65 & 0.7 \\ \cline{2-5}
 & \(\beta\)  & 0.15 & 0.1 & 0.1 \\ \cline{2-5}
 & Learning rate &0.0003 &0.0001 & 0.0003 \\ \cline{2-5}
 & Dropout perc & 0.1 & 0.3 &  0.1 \\ \cline{2-5}
 & Weight decay& ~1.1e-07 & ~0.0005 & ~1.1e-07 \\ \cline{2-5}
 & Num of dense layers & 12 & 4 & 4 \\ \cline{2-5}
 & Size of dense layer & 48 & 32 & 48 \\ \cline{2-5}
 & Batch size & 512 & 2048 &  512 \\ \hline 
{\multirow{8}{*}{LP}} & \(\alpha\) & 0.65 & 0.65 & 0.15 \\ \cline{2-5}
 & \(\beta\)  & 0.55 & 0.45 & 0.15 \\ \cline{2-5}
 & Learning rate & ~4.9e-05 & ~0.002 & ~5.8e-05 \\ \cline{2-5}
 & Dropout perc & 0.2 & 0.3 & 0.2 \\ \cline{2-5}
 & Weight decay & ~8.3e-06 & 3.5e-05 & 7.4e-07 \\ \cline{2-5}
 & Num of dense layers & 14 & 20 & 18 \\ \cline{2-5}
 & Size of dense layer & 64 & 64 & 48 \\ \cline{2-5}
 & Batch size & 1024 & 512 & 512 \\ \hline
\end{tabular}
\label{table:hpt_params_used}
\end{table}

\subsubsection{Inference runtime}

\begin{figure}
\center
\includegraphics[width=0.48\textwidth]{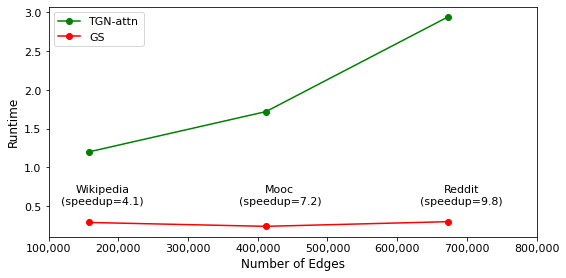}
\caption{\textbf{Speedup vs. number of edges.} The speedups increase almost linearly with the number of edges in the graph.}
\label{fig:run_time_speedup}
\end{figure}

We compare the latency of our framework to baseline GNN architectures. For this purpose, we run 200 batches of 200 events on the external datasets, Wikipedia, Mooc, and Reddit using the node classification task. We compute the average time over 10 runs. Both models were running on Linux PC with 24 Intel Xeon CPU cores (3.70GHz) and a NVIDIA GeForce RTX 2080 Ti GPU (11GB). As depicted in Figure~\ref{fig:run_time_auc}, our graph-sprints consistently outperforms other baselines (TGN-attn, TGN-ID, Jodie) in the node classification task while also demonstrating a significantly lower inference latency. Compared to TGN-attn, the GS achieves better classification results but is close to an order of magnitude faster (Figure~\ref{fig:run_time_speedup}). 

To investigate the impact of graph size on runtime, Figure~\ref{fig:run_time_speedup} showcases our observations. Notably, in the utilized datasets, TGN's runtime increases as the number of edges in the dataset grows, requiring more time to score 200 batches. Conversely, since GS does not require neighborhood sampling, it exhibits constant inference time regardless of the graph size. Furthermore, the speedups achieved by graph-sprints are expected to be significantly higher in a big-data context, where the data is stored in a distributed manner rather than in memory as in our current experiments. In such scenarios, graph operations used in graph-neural networks like TGN-attn would incur even higher computational costs.

Recently, APAN~\cite{wang2021apan} has attempted to build a low-latency framework for CTDGs. Their approach consisted of performing the expensive graph operations asynchronously, out of the inference loop. In that way, they achieved inference speeds of 4.3ms per batch on the Wikipedia dataset, but we cannot directly compare those results with ours (GS: 1.4ms, TGN-attn: 6ms) due to the different setup and hardware. Importantly, their approach achieves low-latency by sacrificing up-to-date information at inference time. Indeed, the inference step is performed without access to the most recent embeddings, because the expensive graph operations to compute the embeddings are performed asynchronously.

\subsubsection{Memory reduction}

Both the Wikipedia and Reddit datasets consist of 172 edge features. By calculating graph-sprints with 10 quantiles per feature, along with incorporating in/out degrees histograms and time-difference histograms, we obtain a node embedding of 1742 features (one feature per histogram bin). In our experimental setup, similar to state-of-the-art approaches, we concatenate the source and destination node embeddings for source label prediction, resulting in a 3484-feature vector. To reduce the size of the node embeddings, we propose a similarity hashing-based memory reduction technique (Section~\ref{sec:reduce_storage})). Our experiments, as presented in Table~\ref{table:results_external_data_memory_reduction}, demonstrate that our technique significantly reduces storage requirements sacrificing the AUC in the node classification task. In the Reddit dataset, storage can be reduced to 50\% with a 0.6\% AUC sacrifice or to 10\% with a 2\% AUC sacrifice. The reduction percentage can be fine-tuned as a hyperparameter, considering the use case and dataset, to strike a balance between precision and memory trade-off.

\begin{table}[h]
\centering
\caption{Effect of memory reduction on the node classification task. We report average test AUC ± std resulting from retraining the best model, after optimization, using 10 random seeds. We use only the GS embeddings to train the classifier. Since Mooc has less features we add '-' where number of features < 1.}

\begin{tabular}{|c|c|c|c|}
\hline
\textbf{Space used} & \textbf{Wikipedia}  & \textbf{Mooc} &  \textbf{Reddit}\\ \hline
 
\textbf{100\%} & 90.7 ± 0.3 & 75.0 ± 0.2 & 68.5 ± 1.0  \\ \hline 
\textbf{50\%} & 90.8 ± 0.1 & 75.0 ± 0.1 & 67.9 ± 1.1 \\ \hline
\textbf{25\%} & 91.1 ± 0.1 & 74.9 ± 0.3 & 65.1 ± 1.9 \\ \hline
\textbf{10\%} & 90.9 ± 0.2  & 74.0  ± 0.3 & 66.5 ± 0.9\\ \hline
\textbf{0.5\%} & 89.7 ± 0.3  & - & 58.0 ± 2.6 \\ \hline
\textbf{0.12\%} & 86.4 ± 0.3  & - & 55.2 ± 1.1 \\ \hline

\end{tabular}
\label{table:results_external_data_memory_reduction}
\end{table}

\subsection{Anti-money laundering (AML) experiments}\label{sec:experiments_aml}
In money laundering, the criminals' objective is to hide the illegal source of their money by moving funds between various accounts and financial institutions. In these experiments, our objective is to enrich a classifier with graph-based features generated by our \textit{graph-sprints} framework.

\subsubsection{Datasets}

We evaluate the graph-sprints framework in the AML domain using two real-world banking datasets. Due to privacy concerns, we can not disclose the identity of the financial institutions (FIs) nor provide exact details regarding the node features. We refer to the datasets as \textit{FI-A} and \textit{FI-B}. The graphs in this use-case are constructed by considering the accounts as nodes and the money transfers between accounts as edges.
Table~\ref{table:internal_data_stats} shows approximate details of these datasets.

\subsubsection{Models and Experimental setup} As before, we train the neural network classifier that uses raw node features only, i.e., no graph information is present (Raw). We compare that baseline performance against models that include only \textit{graph-sprint} features (GS), and models that use both \textit{graph-sprints} features and raw features (GS+Raw). Finally, we train the same GNN architectures as in the public datasets ( TGN-ID, Jodie, and TGN-attn).

\begin{table}[h]
\centering
\caption{Information about AML datasets.}
\begin{tabular}{|c|c|c|}
\hline
\textbf{} & \textbf{FI-A} & \textbf{FI-B} \\ \hline
\#Nodes & $\approx$400000 & $\approx$10000  \\ \hline
\#Edges & $\approx$500000 & $\approx$2000000  \\ \hline
Positive labels & 2-5\% & 20-40\% \\ \hline
Duration &  $\approx$300 days  &  $\approx$600 days \\ \hline
Edges/day (mean ± std)  &  1500 ± 750  & 3000 ± 5000 \\ \hline
Used split &  60\%-10\%-30\%  & 60\%-10\%-30\% \\ \hline
\end{tabular}
\label{table:internal_data_stats}

\end{table}

\subsubsection{Results}

Due to privacy considerations, we are unable to disclose the actual obtained AUC values. Instead, we present the relative improvements in AUC ($\Delta$AUC) when compared to a baseline model that does not utilize graph features. In this context, the baseline model corresponds to a $\Delta$AUC value of 0, and any increase in recall compared to the baselines is represented by positive values of $\Delta$AUC.

Table~\ref{table:aml_results} displays the $\Delta$AUC values for our GS variations and other state-of-the-art baselines. Our GS variations exhibit the most favorable outcomes in both datasets, with an approximate 3.3\% improvement in AUC for the \textit{FI-A dataset} and a 27.8\% improvement in AUC for the \textit{FI-B dataset}.
    
\begin{table}[h]
\centering
\caption{Node classification results in AML data. We report relative gain in AUC ($\Delta$AUC) compared to a baseline model that does not utilize graph features. We report the
average test $\Delta$AUC ± std achieved by retraining the best model
after hyperparameter optimization using 10 random seeds. We identify the \textcolor{red}{\textbf{best}} model and highlight the \textcolor{blue}{\underline{second best}} model.}
\begin{tabular}{|c|c|c|}
\hline
\multirow{2}{*}{\textbf{Method}} &\multicolumn{2}{|c|}{$\Delta$\textbf{AUC ± std}} \\ \cline{2-3}
& \textbf{FI-A} & \textbf{FI-B} \\ \hline
TGN-ID &  +0.1 ± 0.1 & +24.4 ± 0.2  \\ \hline
Jodie &  +0.0 ± 0.1 & +24.5 ± 0.2   \\ \hline

TGN-attn &  +0.3 ± 0.7 & \textcolor{blue}{\underline{+25.1 ± 0.3}}  \\ \hline
GS &  \textcolor{blue}{\underline{+1.8 ± 0.5}} & \textcolor{red}{\textbf{+27.8 ± 0.4}}  \\ \hline
GS+Raw &  \textcolor{red}{\textbf{+3.3 ± 0.3}} & +20.1 ± 3.9  \\ \hline
\end{tabular}
\label{table:aml_results}

\end{table}

\section{Related Work}\label{sec:related_work}

\subsection{Random-walk based methods}

DeepWalk~\cite{perozzi2014deepwalk} and node2vec~\cite{grover2016node2vec} are two high-latency random-walk based methods to extract node embeddings on static graphs. The main limiting factors of these methods are that they disregard node and edge features as well as temporal information, since they are designed for static graphs. \citet{sajjad2019efficient} extend these random-walk based methods to discrete-time dynamic graphs. While some efficiency is gained with the proposed method, it is far from applicable to CTDGs and in low-latency settings. Node2bits~\cite{jin2019node2bits} considers the temporal information by defining several time windows over the sampled random-walks, and aggregates node attributes in these time windows into histograms. Node2bits does not include edge-features and performs costly computations which cannot be performed with low latency.

Continuous-time Dynamic Node Embeddings (CTDNE)~\citep{lee2020dynamic, nguyen2018continuous} were proposed to generate time-aware embeddings, generalizing the node2vec framework to CTDGs. The authors consider the graph as a stream of edges, and propose to perform temporal walks starting from seed nodes chosen from a temporally biased distribution. Similarly, temporal random-walks have been used to extract embeddings into hyperbolic spaces~\citep{wang2021hyperbolic}. Causal anonymous walks~\cite{wang2021inductive} propose to use anonymized walks in order to encode motif information. Similarly, NeurTWs~\cite{jin2022neural} explicitly model time in the anonymous walks using Neural Ordinary Differantial Equations (NeuralODEs). Unlike our graph-sprints framework, full random-walks need to be performed.

\subsection{K-hop neighborhood based methods}
Most GNN-based methods require a K-hop neighborhood on which message-passing operations lead to node embeddings. To deal with CTDGs, a simple approach is to consider a series of discrete snapshots of the graph over time, on which static methods are applied. Such approaches however do not take time properly into account and several works propose techniques to alleviate this issue\cite{sankar2020dysat, goyal2018dyngem, jin2022generalizing, you2022roland}. To better deal with CTDGs, other works focus on including time-aware features or inductive biases into the architecture. DeepCoevolve~\cite{dai2016deep} and Jodie~\cite{kumar2019predicting} train two RNNs for bipartite graphs, one for each node type. Importantly, the previous hidden state of one RNN is also added as an input to the other RNN. In this way, the two RNNs interact, in essence performing single-hop graph aggregations. TGAT~\cite{xu2020inductive} proposes to include temporal information in the form of time encodings, while TGN~\cite{tgn_icml_grl2020} extends this framework and also includes a memory module taking the form of a recurrent neural network.  In~\cite{jin2020static}, the authors replace the discrete-time recurrent network of TGN with a NeuralODE modeling the continuous dynamics of node embeddings. 

APAN~\cite{wang2021apan} proposes to reduce the latency at inference time by decoupling the more costly graph operations from the inference module. The authors propose a more light-weight inference module that computes the predictions based on a node's embedding as well as a node's mailbox, which contains messages of recently interacting nodes. The mailbox is updated asynchronously, i.e. separated from the inference module, and involves the more expensive k-hop message passing. While APAN improves the latency at inference time, it sacrifices some memory since each node's state is now expanded with a mailbox and it potentially uses outdated information at inference time due to asynchronous update of this mailbox. Also towards reducing computational costs of GNNs, HashGNN~\cite{wu2021hashing} leverages MinHash to generate node embeddings suitable for the link prediction task, where nodes that results in the same hashed embedding are considered similar. SGSketch~\cite{yang2022streaming} is a streaming node embedding framework uses a mechanism to gradually forget outdated edges, achieving significant speedups. Differently than our approach SGSketch uses the gradual forgetting strategy to update the adjacency matrix and therefore only considers the graph structure.

\section{Conclusions}
\label{sec:conclusions}

This paper introduces the graph-sprints framework, which enables the computation of time-aware embeddings for CTDGs with minimal latency. The study demonstrates that the graph-sprints features, when combined with a neural network classifier, achieve competitive predictive performance compared to state-of-the-art methods while having a significantly faster inference time, up to approximately an order of magnitude improvement.
Interestingly, our method performs more strongly on the node classification tasks compared to the link prediction tasks. It would be interesting to investigate what features make it powerful on one task, and what is still missing in the other. In future work, it would be interesting to extend the graph-sprints framework to heterogeneous graphs, and explore how GNNs could inherit some of the strengths of graph-sprints.

\balance
\bibliographystyle{ACM-Reference-Format}
\bibliography{06_references}

\end{document}